\documentclass[letterpaper, 10 pt, conference]{ieeeconf}  % Comment this line out if you need a4paper

\IEEEoverridecommandlockouts                              % This command is only needed if 
\usepackage{graphicx,subfig}

\usepackage{hyperref}
\usepackage[sorting=none, style=ieee, url=false]{biblatex} % ✅ sorting=none = 按引用顺序编号
\usepackage{booktabs}  
\usepackage{threeparttable} 
\usepackage{makecell}
\usepackage{graphicx}
\usepackage{multicol}
\usepackage{multirow}
\usepackage{feyn}
\usepackage{amsmath,amssymb,amsfonts}
\usepackage{ulem}
\usepackage{caption}
\usepackage{fancyhdr} % 添加页眉页脚
\usepackage{xcolor} % 在导言区引入
\usepackage{float}
\usepackage{hyperref}

\hypersetup{
	colorlinks=true,
	linkcolor=cyan,
	filecolor=blue,
	urlcolor=black,
	citecolor=green,
}

\title{\LARGE \bf
LiteViLNet: Lightweight Vision-LiDAR Fusion Network for Efficient Road Segmentation
}

\author{Daojie Peng, Bingtao Wang, Fulong Ma, Liang Zhang, Jun Ma$^\dagger$
\thanks{$\dagger$ Corresponding author: \texttt{jun.ma@ust.hk}}
\thanks{Daojie Peng, Fulong Ma and Jun Ma are with The Hong Kong University of Science and Technology (Guangzhou) (e-mail: \{fmaaf, dpeng108\}@connect.hkust-gz.edu.cn, jun.ma@ust.hk.)}
\thanks{Bingtao Wang and Liang Zhang are with The Shandong University, wangbt@mail.sdu.edu.cn, 201299800013@sdu.edu.cn 
}
}

\begin{document}

\maketitle
% \fancyhf{}
% \pagestyle{fancy} %IEEE模板在\maketitle后会自动声明\thispagestyle{plain}，
% %                       % 导致第一页什么都没有。所以得把plain更改为fancy
% \lhead{} % 页眉左，需要东西的话就在{}内添加
% \chead{} % 页眉中
% \rhead{} % 页眉右
% \lfoot{} % 页眉左
% \cfoot{} % 页眉中
% \rfoot{\thepage} %页眉右，\thepage 表示当前页码

\thispagestyle{empty}
\pagestyle{empty}

%%%%%%%%%%%%%%%%%%%%%%%%%%%%%%%%%%%%%%%%%%%%%%%%%%%%%%%%%%%%%%%%%%%%%%%%%%%%%%%%
\begin{abstract}
Road segmentation is a fundamental perception task for autonomous driving and mobile robotics, where both appearance and geometric cues must be processed under edge-computing constraints. Existing multi-modal approaches often improve accuracy with large encoders or expensive global interaction, which limits their use on embedded platforms. We present \textbf{LiteViLNet}, a lightweight RGB-geometry fusion network that combines a MobileNetV3 RGB encoder with a 0.12M-parameter depth-wise-separable geometry encoder. A multi-scale feature fusion module performs modality-specific enhancement, global-query cross-modal interaction, and adaptive gating, while a depth-wise large-kernel bridge enlarges the contextual support of the deepest representation with low overhead. The resulting U-Net-style decoder uses deep supervision only during training. On the KITTI Road benchmark, the 14.04M-parameter full model obtains $97.23\pm0.15\%$ MaxF. On the held-out ORFD test set under the released OFF-Net evaluation protocol, the full model achieves $96.74\pm0.09\%$ F-score and $93.68\pm0.18\%$ IoU. On a Jetson Orin NX, model-only PyTorch FP16 inference reaches $22.18\pm0.21$ FPS; a separate TensorRT FP16 measurement reaches $68.73\pm0.06$ FPS on the Jetson. Camera-depth adaptations and perception-and-control demonstrations on three heterogeneous robot platforms further illustrate the portability of the dual-stream design. 
% \textcolor{blue}{The three-seed ORFD ablation verifies consistent F-score and IoU gains from MSFM and a 0.91-point mean AP gain from the large-kernel bridge.}

\end{abstract}

\section{Introduction}
Drivable area segmentation, which identifies the free road surface from the surrounding environment, is a critical component of autonomous vehicles and intelligent robotic systems, providing a basis for planning and navigation \cite{min2022orfd}.
Recent road-perception research spans RGB-only and single-sensor geometric methods \cite{sun2025rod,milioto2019rangenet++,zhao2025curbnet}, RGB--geometry fusion architectures \cite{sneroad,chen2019progressive,min2022orfd}, and self-supervised or annotation-free learning \cite{ma2023self, ma2026annotation}.
However, deploying high-performance segmentation models on edge devices remains challenging due to strict computation, memory, and power constraints. These constraints have motivated lightweight segmentation and deployment-oriented perception systems in recent years \cite{nadeem2026lightweight,peng2025lovon}.

\begin{figure*}[h]
\centering
\includegraphics[width=1.0\linewidth]{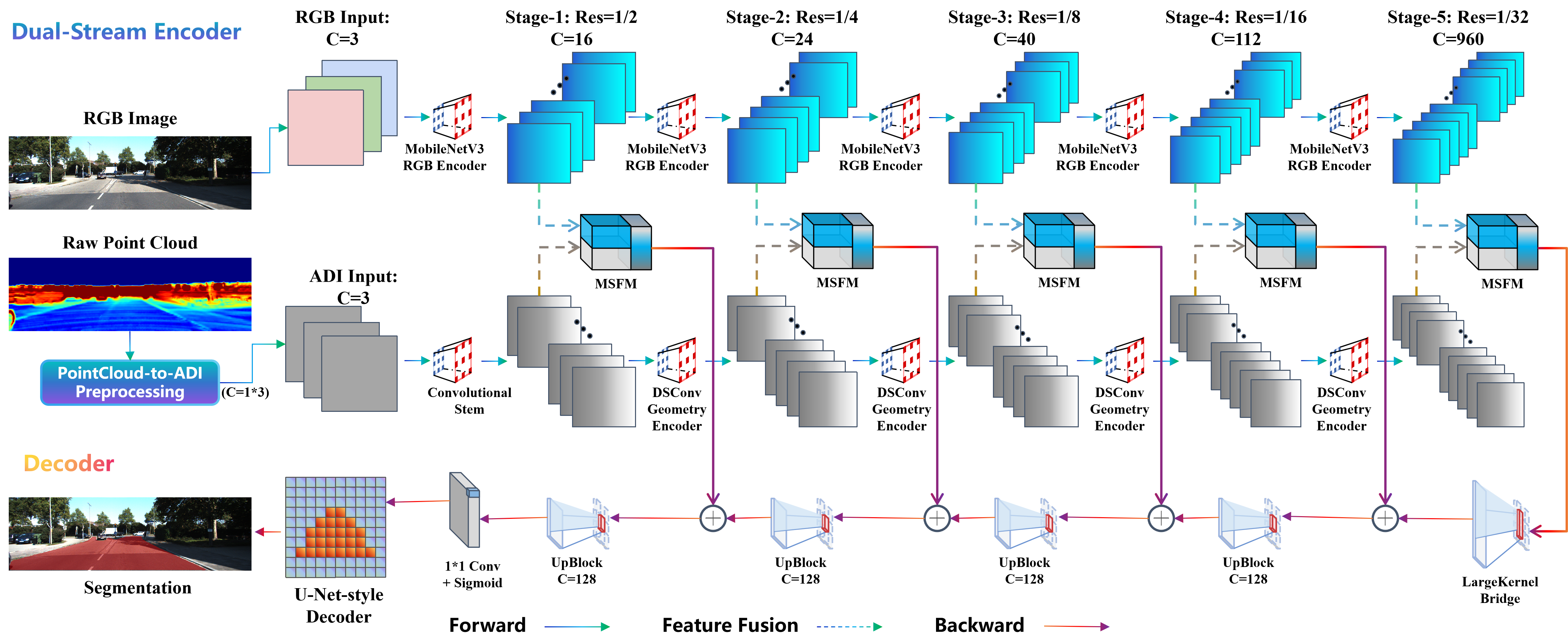}
\caption{\textbf{Overall Architecture of LiteViLNet.} The RGB stream uses MobileNetV3, while the geometry stream employs a \textcolor{black}{0.12M-parameter convolutional stem and four DSConv stages}. For a point-cloud geometric input, \textcolor{black}{PointCloud-to-ADI preprocessing produces a single-channel ADI that is replicated to three channels before the geometry encoder}. The resulting three-channel geometric representation is processed by the same geometry stream across sensing instantiations.}
\label{fig_architecture}
\end{figure*}

Appearance-only road segmentation can become ambiguous under illumination changes and weak visual texture~\cite{cmx2022}, which motivates using geometry as a complementary cue. SNE-RoadSegV2~\cite{snerv2} and RoadFormer~\cite{roadformer}, for example, fuse RGB with normal/depth features. \textcolor{black}{The diverse illumination, weather, and terrain represented in ORFD provide a complementary off-road benchmark for this design choice.}
Additionally, some methods \cite{chen2019progressive,ma2025annotation} pre-process depth point clouds to generate Altitude Difference Images (ADI), which are then used as model inputs. This approach preserves geometric cues in a compact raster representation.
Transformer backbones such as Swin Transformer~\cite{liu2021swin} provide another design option for feature extraction in segmentation networks.
% \textcolor{blue}{LiteViLNet uses compact encoders, global-query cross-modal interaction, and a depth-wise large-kernel bridge; its segmentation accuracy and throughput are evaluated at fixed resolution, backend, precision, and hardware.}

Lightweight segmentation can also be approached through compact RGB-only architectures, distillation, or task-specific multi-modal designs. \textcolor{black}{TwinLiteNet+~\cite{ahmed2025twinlitenet} studies RGB lane/drivable-area prediction, KGD~\cite{li2025knowledge} develops teacher-student training, SDFNet~\cite{wang2025sdfnet} targets urban semantic segmentation, and LCIRE-Net~\cite{zhang2025lcirenet} considers remote-sensing road extraction.} LRDNet~\cite{lrdnet} and USNet~\cite{usnet} are closer RGB-geometry road-segmentation systems.

To this end, we present LiteViLNet, a Lightweight Vision-LiDAR Network for efficient multi-modal road segmentation. The overall architecture is illustrated in Fig. \ref{fig_architecture}. The RGB stream uses a pre-trained MobileNetV3-Large~\cite{howard2019searching}, while a small DSConv encoder processes point-cloud-derived ADI. At each feature scale, the MSFM combines modality-specific enhancement, cross-modal interaction, and adaptive gating; a depth-wise large-kernel bridge enriches the deepest representation without spatial self-attention. 
LiteViLNet couples compact RGB-geometry fusion with deployment evaluation on embedded hardware.
% , rather than in claiming each standard operator as a new primitive.}
\textcolor{black}{More specifically, our contributions are summarized at the architecture and system levels as follows:}
\begin{itemize}

% \item \textcolor{blue}{We develop a compact dual-stream RGB-geometry architecture that couples a standard RGB backbone with a 0.12M-parameter geometry encoder and lightweight multi-scale fusion in a 14.04M-parameter network.}
% \item \textcolor{blue}{We design MSFM with global-query interaction that avoids quadratic spatial attention, together with a depth-wise large-kernel bridge that efficiently expands contextual support.}
% \item \textcolor{black}{We build a deployment-oriented multi-modal perception system that accommodates depth point-cloud inputs from different sensing devices through a common compact geometry stream, while maintaining the same architecture across evaluation and robotic operation.}
\item We introduce \textbf{LiteViLNet}, a compact dual-stream RGB-geometry segmentation architecture that couples a standard RGB backbone with a lightweight 0.12M-parameter geometry encoder and multi-scale cross-modal fusion, yielding an efficient 14.04M-parameter network.
\item We extend the multi‑scale feature fusion module (MSFM) with a global‑query interaction scheme to eliminate the quadratic complexity of spatial attention, and further propose a depth‑wise large‑kernel bridge (LKB) to efficiently enlarge the contextual receptive field.
\item We build a deployment-oriented multi-modal perception system whose common compact geometry stream accommodates point-cloud inputs from different sensing devices, while keeping the same architecture across offline evaluation and on-robot operation. 
\item We conduct extensive experiments on the KITTI/ORFD datasets and real-world robotic deployments (including Kuafu Delivery Vehicle, Unitree-B2, and Unitree G1). The results demonstrate that LiteViLNet achieves
promising performance with only 14.04M parameters and 22.18
FPS model-only inference speed on Jetson, significantly outperforming
existing lightweight methods in speed while maintaining competitive accuracy.

\end{itemize}

\section{Related Works}

\subsection{Road Segmentation}
Road segmentation has been extensively studied over the past decades. Early works primarily rely on hand-crafted features, such as color and texture \cite{early_road}. With the rise of deep learning, Fully Convolutional Networks (FCNs) \cite{long2015fully} and encoder-decoder architectures like U-Net \cite{ronneberger2015u} have become the standard paradigm. Recent works have further improved the performance by introducing more powerful backbones. For example, SNE-RoadSeg \cite{sneroad} introduces surface normal estimation to help road segmentation. Later, SNE-RoadSegV2 \cite{snerv2} extended this work by using Swin Transformer as the backbone. RoadFormer \cite{roadformer} further proposes a transformer-based fusion framework to model the cross-modal interactions. 
LiteViLNet advances geometry-assisted road segmentation with a 0.12M-parameter geometry encoder, global-query fusion, and hardware-verified edge throughput.

\subsection{Multi-Modal Fusion for Segmentation}
Multi-modal fusion combines RGB with depth, LiDAR, thermal, or other complementary measurements. Early fusion methods concatenate input channels \cite{early_fusion}, whereas CMX \cite{cmx2022} calibrates cross-modal features. In a related camera-only setting, Cross-View Transformers \cite{crossview} model longer-range dependencies across camera views for map-view semantic segmentation.
LiteViLNet extends structured cross-modal interaction to a compact dual-stream network, with segmentation accuracy, trainable parameters, MAC-equivalent cost, and measured latency reported on embedded hardware.

\subsection{Lightweight Semantic Segmentation}
To enable deployment on edge devices, MobileNet~\cite{howard2017mobilenets}, ShuffleNet~\cite{zhang2018shufflenet}, ESPNet~\cite{mehta2018espnet}, and BiSeNet~\cite{yu2018bisenet} reduce segmentation cost through compact operators and backbones. For geometry-assisted road segmentation, LRDNet~\cite{lrdnet} uses LiDAR and USNet~\cite{usnet} uses RGB with depth-derived surface-normal input.
LiteViLNet brings these efficiency principles to RGB-geometry road segmentation through global-query fusion and a depth-wise large-kernel bridge.

\begin{figure*}[h]
    \centering
    \includegraphics[width=0.6\linewidth]{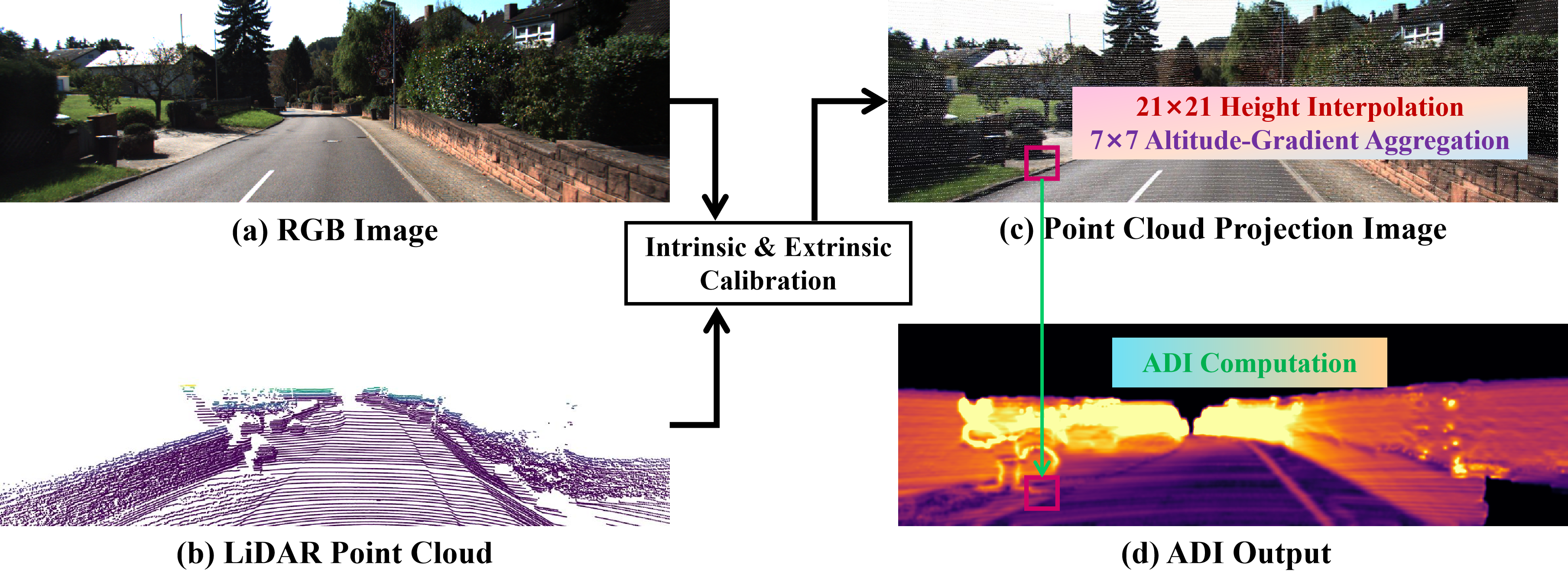}
    \caption{\textbf{\textcolor{black}{PointCloud-to-ADI preprocessing used for the KITTI point-cloud input.}} Calibrated projection first produces a sparse height map; inverse-distance interpolation in a \textcolor{black}{$21\times21$ window}, local altitude-gradient aggregation in a \textcolor{black}{$7\times7$ window}, and per-image normalization/Gaussian smoothing then produce the ADI. This preprocessing is performed before the network, and the resulting three-channel geometric representation is fed to the common geometry encoder.}
    \label{fig:adi}
\end{figure*}

\section{Method}
In this section, we present the details of the \textcolor{black}{implemented LiteViLNet framework}. The overall architecture is illustrated in Fig. \ref{fig_architecture}.

\subsection{Input Representation}\label{sec:input}
The input representation is demonstrated in Fig. \ref{fig:adi}. Given an RGB image $I\in\mathbb{R}^{H\times W\times3}$ and a depth point cloud, the KITTI instantiation consumes \textcolor{black}{precomputed ADI files}; reproducible regeneration follows the \textcolor{black}{released PLARD reference procedure}~\cite{chen2019progressive}. For KITTI, the point cloud is supplied by LiDAR. A point $X_i=[x_i,y_i,z_i,1]^\top$ with \textcolor{black}{$x_i\geq5$ m} is projected by \textcolor{black}{$P_2R_{0,\mathrm{rect}}T_{\mathrm{velo}\rightarrow\mathrm{cam}}X_i$} and rounded to an image pixel. Out-of-frame points are discarded; when several points reach one pixel, \textcolor{black}{the point with the smallest Euclidean range is retained}. This gives a sparse height image $z_p$ and validity mask $m_p$.

For every unobserved pixel $p$, height is interpolated from valid pixels in the \textcolor{black}{$21\times21$ neighborhood $\mathcal N_{10}(p)$ using $w_{pq}=1/\max(\lVert p-q\rVert_2,1)$}. The weighted interpolation is defined as
\begin{align}
\textcolor{black}{Z_p}&\textcolor{black}{=\sum_{q\in\mathcal N_{10}(p)}m_qw_{pq},}\nonumber\\
\textcolor{black}{\widetilde z_p}&\textcolor{black}{=
\begin{cases}
z_p, & m_p=1,\\
Z_p^{-1}\!\displaystyle\sum_{q\in\mathcal N_{10}(p)}m_qw_{pq}z_q,
& m_p=0,\ Z_p>0,\\
0, & m_p=0,\ Z_p=0.
\end{cases}}
\label{eq:adi_interpolation}
\end{align}
Here, \textcolor{black}{$K_{\mathrm{int}}=21$ and the updated validity is $\widetilde m_p=\mathbf 1[m_p=1\ \text{or}\ Z_p>0]$}; hence, an unobserved pixel with no valid neighbor remains invalid. For every valid pixel after interpolation, the altitude-gradient magnitude is then averaged over valid pixels in the \textcolor{black}{$7\times7$ neighborhood $\mathcal N_3(p)$}:
\begin{align}
\textcolor{black}{\mathcal V_p}&\textcolor{black}{=\{q\in\mathcal N_3(p):\widetilde m_q=1\},}\nonumber\\
\textcolor{black}{a_p}&\textcolor{black}{=\frac{1}{|\mathcal V_p|}\sum_{q\in\mathcal V_p}
\sqrt{g_x(p,q)^2+g_y(p,q)^2},}
\label{eq:adi_gradient}\\[-1mm]
\textcolor{black}{g_x(p,q)}&\textcolor{black}{=
\begin{cases}(\widetilde z_q-\widetilde z_p)/(q_x-p_x),&q_x\neq p_x,\\0,&q_x=p_x,\end{cases}}\nonumber\\
\textcolor{black}{g_y(p,q)}&\textcolor{black}{=
\begin{cases}(\widetilde z_q-\widetilde z_p)/(q_y-p_y),&q_y\neq p_y,\\0,&q_y=p_y.\end{cases}}\nonumber
\end{align}
The interpolation and gradient stages therefore use distinct neighborhood sizes, \textcolor{black}{$K_{\mathrm{int}}=21$ and $K_{\mathrm{grad}}=7$}, respectively. The resulting gradient map is normalized independently for each image: if \textcolor{black}{$\mathcal P=\{p:a_p>0\}$} is nonempty, we set \textcolor{black}{$b_p=20(a_p-\min_{q\in\mathcal P}a_q)$ for $p\in\mathcal P$ and $b_p=0$ otherwise}, and obtain \textcolor{black}{$A=\operatorname{clip}_{[0,1]}(G_{\sigma=2}*\sqrt{b})$}. If $\mathcal P$ is empty, the procedure returns an all-zero ADI. Finally, $A$ is replicated to three channels and fixed per-channel normalization is applied.

\subsection{Dual-Stream Lightweight Encoder}
We use a dual-stream architecture to extract features from the RGB and depth-point-cloud geometry modalities separately.
\begin{itemize}
\item \textbf{RGB Stream}: We adopt ImageNet-pretrained MobileNetV3-Large \cite{howard2019searching} as the RGB backbone to obtain an efficient multi-scale feature hierarchy. It consists of five stages, generating a feature pyramid ${F^{rgb}_l}_{l=0}^{4}$ with resolutions of $1/2, 1/4, 1/8, 1/16, 1/32$ of the input size, respectively.
\item \textbf{Geometry Stream}: The geometry stream uses \textcolor{black}{a small two-convolution stem followed by four depth-wise-separable convolution (DSConv) stages}. Its five outputs align with the RGB stream at \textcolor{black}{spatial resolutions $(1/2,1/4,1/8,1/16,1/32)$ and channel widths $[16,24,40,112,960]$}, enabling efficient scale-wise fusion with only 0.12M trainable parameters.
\end{itemize}

\subsection{Multi-Scale Feature Fusion Module}\label{sec:msfm}

\begin{figure}[t]
    \centering
    \includegraphics[width=0.98\linewidth]{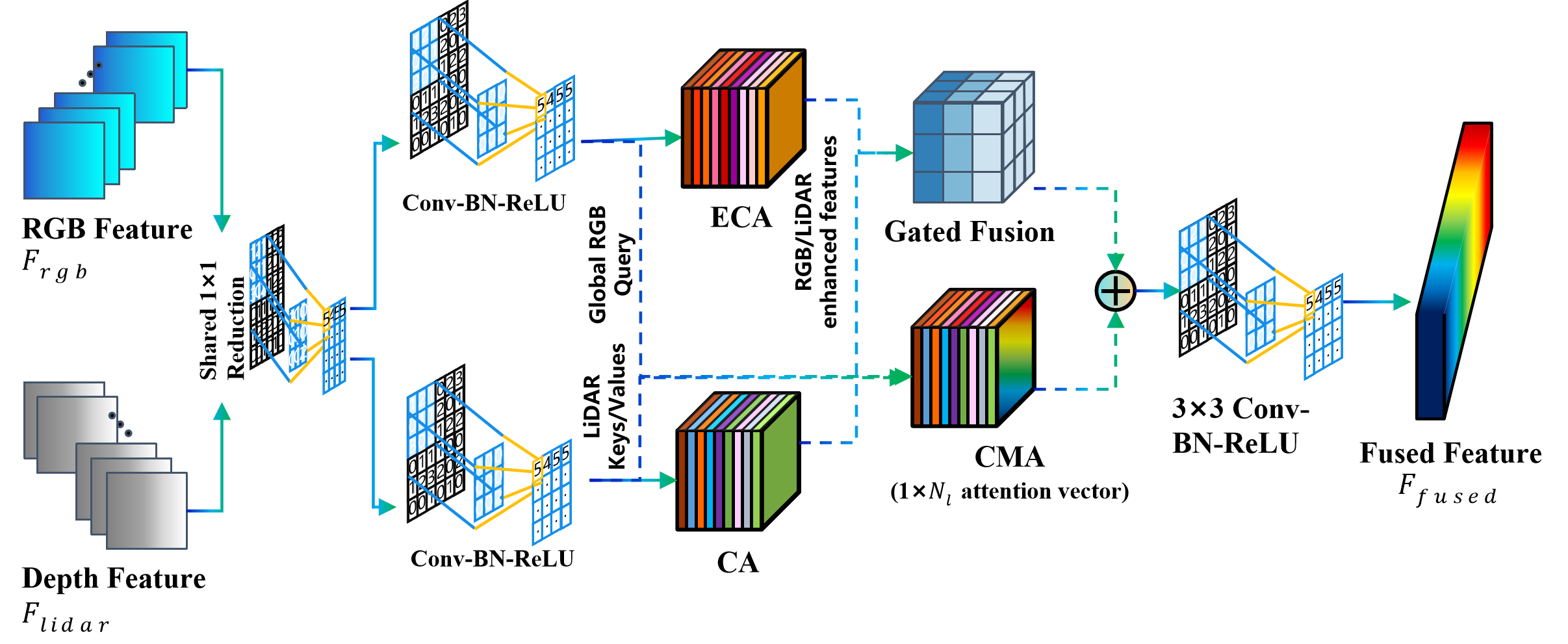}
    \caption{\textbf{MSFM at one feature scale.} The implemented CMA uses \textcolor{black}{one globally pooled RGB query and spatial geometry keys/values}, producing a \textcolor{black}{$1\times N_l$ attention vector}. The gate is computed from the two enhanced modality features.}
    \label{fig_msfm}
\end{figure}

To fuse the two modalities, we design a \textcolor{black}{Multi-Scale Feature Fusion Module (MSFM) that combines modality-specific enhancement, global-query cross-modal attention, and adaptive gating}. As shown in Fig. \ref{fig_msfm}, the module is applied at each of the five scales.
For each scale $l$, given the RGB feature $F^{rgb}_l \in \mathbb{R}^{B \times C_l \times H_l \times W_l}$ and geometry feature $F^{geo}_l \in \mathbb{R}^{B \times C_l \times H_l \times W_l}$, the fusion process is defined as follows:
\begin{enumerate}
\item \textbf{Dimension Reduction}: We first use a $1 \times 1$ convolution to reduce the channel dimension of both features from $C_l$ to $C_l/2$, reducing the computational cost of the subsequent attention calculation:
\begin{align}
    F^{rgb'}_l = \text{Conv}_{1\times1}\left(F^{rgb}_l\right),\nonumber\\
    F^{geo'}_l = \text{Conv}_{1\times1}\left(F^{geo}_l\right)\nonumber
\end{align}
\textcolor{black}{The implementation uses the same reduction layer for the two modalities at a given scale.}

\item \textbf{Intra-modal Enhancement}: \textcolor{black}{A $3\times3$ Conv-BN-ReLU block is applied to each dimension-reduced feature.} Efficient Channel Attention (ECA)~\cite{eca} then reweights the RGB channels, while Coordinate Attention (CA)~\cite{coordinate} supplies direction-aware spatial reweighting for the geometry feature:
\begin{align}
    \textcolor{black}{\hat{F}^{rgb}_l = \text{ECA}\left(\phi^{rgb}_{3\times3}(F^{rgb'}_l)\right),}\nonumber\\
    \textcolor{black}{\hat{F}^{geo}_l = \text{CA}\left(\phi^{geo}_{3\times3}(F^{geo'}_l)\right).}\nonumber
\end{align}

\item \textbf{Cross-modal Attention (CMA)}:
The implementation uses \textcolor{black}{a single global RGB query to select informative locations from the geometry feature}. Let $d=C_l/2$, $N_l=H_lW_l$, and $\operatorname{vec}(\cdot)$ flatten only the spatial axes. The exact operations are
\begin{align}
\textcolor{black}{Q_l}&\textcolor{black}{=W_q\operatorname{GAP}(\hat F^{rgb}_l)
\in\mathbb R^{B\times1\times d},}\nonumber\\
\textcolor{black}{K_l}&\textcolor{black}{=\operatorname{vec}(W_k\hat F^{geo}_l)
\in\mathbb R^{B\times N_l\times d},}\nonumber\\
\textcolor{black}{V_l}&\textcolor{black}{=\operatorname{vec}(W_v\hat F^{geo}_l)
\in\mathbb R^{B\times d\times N_l},}\nonumber\\
\textcolor{black}{\alpha_l}&\textcolor{black}{=\operatorname{softmax}\!\left(Q_lK_l^\top/\sqrt d\right)
\in\mathbb R^{B\times1\times N_l},}
\label{eq:cma_linear}\\
\textcolor{black}{c_l}&\textcolor{black}{=\operatorname{LN}\!\left(W_o(V_l\alpha_l^\top)\right),}\nonumber\\
\textcolor{black}{F^{cross}_l}&\textcolor{black}{=\hat F^{rgb}_l+\operatorname{Expand}_{H_l,W_l}(c_l).}\nonumber
\end{align}
Thus, only $B N_l$ attention scores are materialized. For fixed $d$, the attention aggregation has $O(BN_ld)$ arithmetic and $O(BN_l)$ score memory, rather than $O(BN_l^2)$. At the highest-resolution scale for a $384\times1248$ input, $N_0=192\times624=119{,}808$; its FP16 score vector occupies approximately 0.23~MiB per sample.

\item \textbf{Gated Fusion}: The gate is generated from the \textcolor{black}{concatenated enhanced RGB and geometry features}. It balances these two features; \textcolor{black}{the CMA residual is then added, and a $3\times3$ Conv-BN-ReLU block restores $C_l$ channels}:
\begin{align}
    \textcolor{black}{g_l=\sigma\!\left(\operatorname{BN}\!\left(W_g[\hat F^{rgb}_l;\hat F^{geo}_l]\right)\right),}\nonumber\\
    \textcolor{black}{M_l=g_l\odot\hat F^{rgb}_l+(1-g_l)\odot\hat F^{geo}_l,}\nonumber\\
    \textcolor{black}{F^{fused}_l=\phi_{3\times3}\!\left(M_l+F^{cross}_l\right).}\nonumber
\end{align}
\end{enumerate}
The five scale-specific modules add \textcolor{black}{10.35M parameters} and enrich the fused representation at every encoder resolution.

\subsection{Large-Kernel-Bridge Semantic Enhancement}\label{sec:bridge}
After fusion, the deepest feature $F^{fused}_4$ has the highest semantic level and $1/32$ input resolution. We configure a bottleneck large-kernel bridge, using a $7\times7$ depth-wise convolution to broaden contextual support without the parameter and MAC overhead of spatial self-attention.

The module adopts a bottleneck structure to control the parameters, and the forward process is defined as:
\begin{align}
    \textcolor{black}{U=\operatorname{GELU}\!\left(\operatorname{BN}\!\left(W_{down}^{1\times1}F^{fused}_4\right)\right),}\nonumber\\
    \textcolor{black}{V=\operatorname{Drop}_{0.2}\!\left(\operatorname{GELU}\!\left(\operatorname{BN}\!\left(\operatorname{DWConv}_{7\times7}(U)\right)\right)\right),}\nonumber\\
    \textcolor{black}{F^{enh}_4=F^{fused}_4+\operatorname{BN}\!\left(W_{up}^{1\times1}V\right).}
\end{align}
$W_{down}^{1\times1}$ reduces 960 channels to 128 and $W_{up}^{1\times1}$ restores 960 channels. The bridge adds \textcolor{black}{0.254M parameters}. 
% \textcolor{blue}{Its incremental effect is evaluated within the cumulative ORFD ablation in Section~\ref{sec:ablation}.}

\subsection{Decoder and Deep Supervision}
We use a U-Net style decoder to recover the spatial resolution. The decoder takes the enhanced deepest feature as input and gradually upsamples it, concatenating with the fused features from the encoder via skip connections. The decoder consists of one bottleneck layer and four UpBlocks, with channel configuration [128, 64, 32, 16, 16].

The bottleneck layer first reduces the channel dimension of $F^{enh}_4$ from 960 to 128 via a $1 \times 1$ convolution. Then, each UpBlock performs upsampling and feature fusion sequentially, defined as:
\[
D_l = \text{DoubleConv}\left(\left[\text{Up}\left(D_{l+1}\right); \text{Conv}_{1\times1}\left(F^{fused}_l\right)\right]\right)
\]
where $\text{Up}(\cdot)$ is the bilinear upsampling to match the spatial resolution of the skip connection feature, $\text{Conv}_{1\times1}$ adjusts the channel dimension of the skip connection feature to match the current decoder layer, $[\cdot;\cdot]$ is channel-wise concatenation, and $\text{DoubleConv}$ consists of two cascaded $3\times3$ convolution-BatchNorm-ReLU blocks.

During training, three auxiliary segmentation heads are evaluated after decoder blocks with 64, 32, and 16 channels. During inference \textcolor{black}{their forward calls are bypassed; their 115 stored parameters remain part of the reported parameter count but incur no inference MACs}.

The total loss function is a combination of the main loss and the auxiliary losses:
\[
\mathcal{L}_{total} = \mathcal{L}_{main}(\hat{Y}, Y) + \sum_{k=1}^{3} w_k \cdot \mathcal{L}_{aux}(\hat{Y}^{aux}_k, Y)
\]
where \textcolor{black}{each auxiliary logit is bilinearly resized to the ground-truth resolution inside the loss implementation} and $w_k=[0.4,0.3,0.2]$ for the three heads.

The source implementation uses the same compound loss for the main and auxiliary predictions:
\[
\textcolor{black}{\mathcal L_{seg}=\mathcal L_{BCE}+\mathcal L_{Dice}+0.1\mathcal L_{boundary}.}
\]
$\mathcal L_{boundary}$ is the mean-squared difference between prediction and target boundary responses, each computed as the absolute residual from $3\times3$ average pooling. \textcolor{black}{The compound objective jointly promotes region coverage, class-balanced overlap, and boundary alignment.}

\begin{table*}[h]
\centering
\caption{\textbf{Matched-protocol KITTI perspective-view accuracy and local model-only efficiency.} \textcolor{black}{All accuracy rows use the same category-stratified training/validation partition, $384\times1248$ input, 150-epoch budget, and pixel-accumulated 101-threshold evaluator; values are mean $\pm$ sample SD over $n=3$ runs.} Params are trainable parameters. FPS-1 uses $384\times1248$, batch-1 PyTorch FP16 on Jetson Orin NX and is averaged over three repeated model-only measurements. It excludes preprocessing, transfer, and postprocessing. ``-'' denotes an unmeasured target-device timing.}
\resizebox{0.85\linewidth}{!}{
\begin{tabular}{llcccccc}
\toprule
Method & Input & $n$ & MaxF (\%) & PRE (\%) & REC (\%) & Params (M) & FPS-1\\
\midrule
USNet \cite{usnet} & \textcolor{black}{RGB+Normal} & \textcolor{black}{3} & \textcolor{black}{$\mathbf{97.88{\pm}0.07}$} & \textcolor{black}{$\mathbf{98.03{\pm}0.03}$} & \textcolor{black}{$\mathbf{97.73{\pm}0.11}$} & 30.74 & \textcolor{black}{6.03}\\
SNE-RoadSeg \cite{sneroad} & RGB+Normal & \textcolor{black}{3} & \textcolor{black}{$97.23{\pm}0.21$} & \textcolor{black}{$97.39{\pm}0.30$} & \textcolor{black}{$97.06{\pm}0.13$} & 201.32 & \textcolor{black}{2.24}\\
PLARD \cite{chen2019progressive} & RGB+ADI & \textcolor{black}{3} & \textcolor{black}{$95.25{\pm}0.19$} & \textcolor{black}{$95.46{\pm}0.29$} & \textcolor{black}{$95.03{\pm}0.09$} & \textcolor{black}{76.93} & \textcolor{black}{3.38}\\
RoadFormer \cite{roadformer} & RGB+Normal & \textcolor{black}{3} & \textcolor{black}{$97.28{\pm}0.05$} & \textcolor{black}{$97.96{\pm}0.09$} & \textcolor{black}{$96.61{\pm}0.16$} & \textcolor{black}{206.86} & -\\
\textcolor{black}{OFF-Net \cite{min2022orfd}} & \textcolor{black}{RGB+Normal} & \textcolor{black}{3} & \textcolor{black}{$95.36{\pm}0.66$} & \textcolor{black}{$94.88{\pm}0.92$} & \textcolor{black}{$95.83{\pm}0.57$} & \textcolor{black}{25.21} & \textcolor{black}{6.60}\\
\textbf{LiteViLNet (Ours)} & RGB+ADI & \textcolor{black}{3} & \textcolor{black}{$97.23{\pm}0.15$} & \textcolor{black}{$97.31{\pm}0.59$} & \textcolor{black}{$97.16{\pm}0.30$} & \textbf{14.04} & \textcolor{black}{$\mathbf{22.18{\pm}0.21}$}\\
\bottomrule
\end{tabular}
}
\label{tab_compare}
\end{table*}

\section{Experiments}
\subsection{Experimental Setup}\label{sec:setup}
\noindent\textbf{Datasets and protocols}: KITTI Road~\cite{kitti} provides 289 labeled RGB-LiDAR training images and a separate 290-image server test set. \textcolor{black}{We use a category-stratified perspective-view split with 231 training and 58 validation images; the validation set is used for model selection and threshold analysis, not for training. We additionally use the official ORFD~\cite{min2022orfd} training, validation, and testing partitions (8,392/1,245/2,193 image pairs), which cover diverse off-road terrain, weather, and illumination. ORFD registered dense depth is encoded as \texttt{depth3}.}

\noindent\textbf{Metrics}: For KITTI, \textcolor{black}{pixel counts are accumulated over the complete perspective-view validation partition} while probabilities are swept over 101 thresholds; we report MaxF, PRE, REC, AP, and IoU. For ORFD, we additionally follow the released OFF-Net test evaluator: \textcolor{black}{a fixed argmax/0.5 prediction is resized by nearest neighbor to the original $1280\times720$ ground truth and one confusion matrix yields F-score, PRE, REC, and IoU}. AP is reported for our models as a complementary threshold-swept ranking metric. Efficiency reporting includes trainable parameters, fvcore-supported MAC-equivalent cost, and repeated model-only latency.

\noindent\textbf{Implementation and repetition}: We use ImageNet initialization for MobileNetV3 and AdamW with an initial learning rate of $2\times10^{-4}$ and weight decay of $5\times10^{-4}$, followed by cosine annealing. \textcolor{black}{Training uses mixed precision and deterministic cuDNN on RTX 4090~D GPUs. KITTI and ORFD inputs are resized to $384\times1248$ and $704\times1280$, respectively. We report mean and sample standard deviation over three independently initialized runs}.
For the direct comparison in Table~\ref{tab_compare}, USNet, SNE-RoadSeg, PLARD, RoadFormer, and OFF-Net are retrained from the authors' official implementations using the same KITTI split, input resolution, 150-epoch budget, and evaluator as LiteViLNet, while retaining each baseline's released architecture and method-specific training recipe. 
% Because the baselines use their native RGB+normal or RGB+ADI inputs, the table compares complete systems under a common protocol rather than identical sensing modalities. 
Table~\ref{tab_orfd} follows the same principle on the official ORFD partitions: locally trained methods use three independent runs with validation-based checkpoint selection, and all rows are evaluated under the released held-out-test convention. PLARD is not adapted because its official input path requires a LiDAR-derived ADI that is not released for ORFD.

\noindent\textbf{Latency scope}: All latency and throughput tests were conducted on an NVIDIA Jetson Orin NX. Table~\ref{tab_compare} reports FPS-1 using batch 1, $384\times1248$, and PyTorch FP16, with three repeated measurements. Preprocessing, host-to-device transfer, postprocessing, and robot control are excluded from model-only timing. The Jetson uses the maximum-performance \texttt{MAXN\_SUPER} power mode. PyTorch and TensorRT are reported as separate backends.

% \subsection{Comparison with State-of-the-Art Methods}\label{sec:comparison}

% \textcolor{blue}{Table~\ref{tab_compare} provides a directly comparable three-run evaluation under one local perspective-view protocol. LiteViLNet matches SNE-RoadSeg at $97.23\%$ mean MaxF while using 93.0\% fewer parameters. Under the common Jetson PyTorch FP16 protocol, its $22.18\pm0.21$ FPS corresponds to $3.68\times$, $9.88\times$, $6.56\times$, and $3.36\times$ the throughput of USNet, SNE-RoadSeg, PLARD, and OFF-Net, respectively. Its 14.04M parameters and 10.17 GMAC-eq are also substantially below USNet's 30.74M parameters and 39.13 GMAC-eq.}

% \textcolor{blue}{The ORFD comparison in Table~\ref{tab_orfd} extends the analysis beyond KITTI's compact urban-road benchmark to 2,193 held-out frames drawn from 8,392 training and 1,245 validation pairs, with woodland, farmland, grassland, countryside, diverse weather, and challenging illumination. LiteViLNet full reaches $96.74\pm0.09\%$ F-score, $98.31\pm0.37\%$ AP, and $93.68\pm0.18\%$ IoU, exceeding the three-run USNet and SNE-RoadSeg references by 0.79/1.46 and 3.17/5.76 points in F-score/IoU, respectively, and the released OFF-Net checkpoint by 3.94/7.11 points. LiteViLNet therefore leads the evaluated methods in F-score, AP, and IoU on this heterogeneous off-road benchmark.}

% The qualitative results in Fig.~\ref{fig_qualitative} show that LiteViLNet preserves coherent road regions and fine boundary structure across the UM, UMM, and UU scene categories.

\subsection{Comparison with State-of-the-Art Methods}\label{sec:comparison}

\begin{figure*}[h]
\centering
\includegraphics[width=0.90\linewidth]{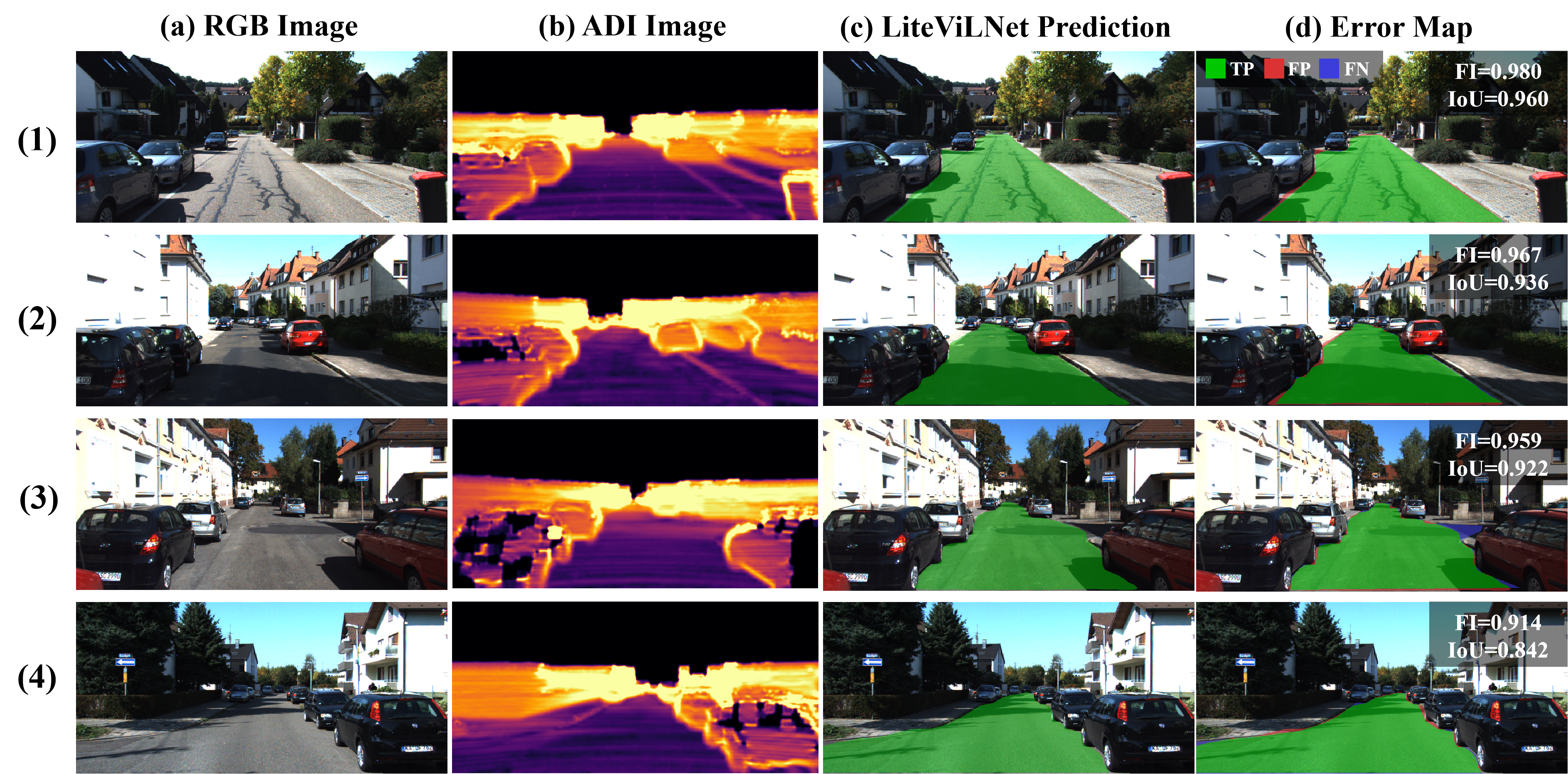}
\caption{\textbf{
% Qualitative KITTI results.
Qualitative Segmentation Results on the KITTI Road Validation Set.
} 
% Each row shows (a) RGB, (b) its corresponding stored ADI, (c) the full-model prediction, and (d) TP/FP/FN in green/red/blue. The examples are selected from the validation set. 
Each row shows (a) the input RGB image, (b) the corresponding Altitude Difference Image (ADI) derived from LiDAR depth data, (c) the segmentation prediction of LiteViLNet, and (d) the error map visualizing true positives (TP, green), false positives (FP, red), and false negatives (FN, blue).
Quantitative metrics including F1-score and IoU are reported for each example. Our method can accurately segment the road area in various scenarios, with errors mainly concentrated on the boundaries.
}
\label{fig_qualitative}
\end{figure*}

Table~\ref{tab_compare} presents a directly comparable three‑run evaluation under a unified local perspective‑view protocol. LiteViLNet achieves a mean MaxF of $97.23\%$, matching the performance of SNE‑RoadSeg while utilizing 93.0\% fewer parameters. Evaluated on Jetson with PyTorch FP16 inference, LiteViLNet runs at $22.18\pm0.21$ FPS, delivering $3.68\times$, $9.88\times$, $6.56\times$, and $3.36\times$ higher throughput compared with USNet, SNE‑RoadSeg, PLARD, and OFF‑Net, respectively. Its total parameter count of 14.04M is also considerably lower than USNet’s 30.74M parameters. The ORFD comparison in Table~\ref{tab_orfd} extends the evaluation beyond KITTI’s constrained urban‑road scenes, covering 8,392 training, 1,245 validation, and 2,193 held‑out test frames with woodland, farmland, grassland, countryside, varied weather and challenging illumination conditions. The full LiteViLNet model obtains $96.74\pm0.09\%$ F‑score, $98.31\pm0.37\%$ AP, and $93.68\pm0.18\%$ IoU. It outperforms the three‑run USNet and SNE‑RoadSeg baselines by 0.79/1.46 and 3.17/5.76 points in F‑score/IoU respectively, and surpasses the released OFF‑Net checkpoint by 3.94/7.11 points, attaining leading F‑score, AP, and IoU among evaluated approaches on this heterogeneous off‑road benchmark. Qualitative results in Fig.~\ref{fig_qualitative} further show that LiteViLNet maintains coherent road regions and fine boundary details across the UM, UMM, and UU scene categories, confirming its favorable accuracy‑efficiency trade‑off for both urban driving and complex off‑road robotic perception.

\begin{table*}[t]
\centering
\caption{\textcolor{black}{\textbf{ORFD held-out testing at $704\times1280$.} F/PRE/REC/IoU follow the released OFF-Net fixed-argmax evaluator at the original $1280\times720$ GT resolution. AP is threshold-swept on the common input grid. USNet, SNE-RoadSeg, and LiteViLNet report mean $\pm$ sample SD over three runs; the OFF-Net released checkpoint is a single local evaluation under the same held-out test convention. Validation is used only for checkpoint selection. All metrics are percentages.}}
\resizebox{\linewidth}{!}{
\begin{tabular}{lcccccccc}
\toprule
Variant & Input & Params (M) & $n$ & F-score & AP & PRE & REC & IoU \\
\midrule
\textcolor{black}{USNet~\cite{usnet}} & \textcolor{black}{RGB+Normal} & \textcolor{black}{30.74} & \textcolor{black}{3} & \textcolor{black}{$95.95{\pm}0.61$} & \textcolor{black}{$97.81{\pm}0.60$} & \textcolor{black}{$95.54{\pm}1.22$} & \textcolor{black}{$96.36{\pm}0.41$} & \textcolor{black}{$92.21{\pm}1.13$} \\
\textcolor{black}{SNE-RoadSeg~\cite{sneroad}} & \textcolor{black}{RGB+Normal} & \textcolor{black}{201.32} & \textcolor{black}{3} & \textcolor{black}{$93.57{\pm}0.77$} & \textcolor{black}{$96.48{\pm}1.75$} & \textcolor{black}{$90.96{\pm}1.63$} & \textcolor{black}{$96.34{\pm}0.22$} & \textcolor{black}{$87.92{\pm}1.35$} \\
\textcolor{black}{OFF-Net~\cite{min2022orfd} released checkpoint} & \textcolor{black}{RGB+Normal} & \textcolor{black}{25.21} & \textcolor{black}{1} & \textcolor{black}{92.80} & \textcolor{black}{97.58} & \textcolor{black}{94.53} & \textcolor{black}{91.13} & \textcolor{black}{86.57} \\
\midrule
\textbf{LiteViLNet full (RGB+geometry)} & \textcolor{black}{RGB+Geometry} & \textcolor{black}{\textbf{14.04}} & 3 & \textcolor{black}{$\mathbf{96.74{\pm}0.09}$} & \textcolor{black}{$\mathbf{98.31{\pm}0.37}$} & \textcolor{black}{$\mathbf{97.03{\pm}0.44}$} & \textcolor{black}{$\mathbf{96.45{\pm}0.39}$} & \textcolor{black}{$\mathbf{93.68{\pm}0.18}$} \\
\bottomrule
\end{tabular}}
\label{tab_orfd}
\end{table*}

\begin{table*}[t]
\centering
\caption{\textbf{Three-run cumulative ablation on the ORFD held-out test set at $704\times1280$.} F-score and IoU follow the released OFF-Net fixed-argmax evaluator at the original $1280\times720$ ground-truth resolution, while AP is threshold-swept on the common input grid. All metrics are percentages and are reported as mean $\pm$ sample SD over three independently initialized runs. Simple denotes scale-wise element-wise addition, and DS denotes deep supervision during training.}
\resizebox{0.8\linewidth}{!}{
\color{black}\begin{tabular}{lccccccc}
\toprule
Variant & Fusion & Bridge & DS & $n$ & F-score & AP & IoU \\
\midrule
% Simple-fusion control & Simple & LKB & \checkmark & 3 & $96.77{\pm}0.27$ & $97.60{\pm}0.41$ & $93.74{\pm}0.52$ \\
% \midrule
RGB-only & - & - & - & 3 & $96.33{\pm}0.18$ & $97.56{\pm}0.46$ & $92.91{\pm}0.34$ \\
+ Geometry & Simple & - & - & 3 & $96.01{\pm}0.90$ & $97.84{\pm}0.30$ & $92.34{\pm}1.66$ \\
+ MSFM & MSFM & - & - & 3 & $96.75{\pm}0.38$ & $97.53{\pm}0.77$ & $93.71{\pm}0.72$ \\
+ Large-kernel bridge & MSFM & LKB & - & 3 & $96.83{\pm}0.55$ & $98.44{\pm}0.39$ & $93.85{\pm}1.02$ \\
\midrule
\textbf{Full} & MSFM & LKB & \checkmark & 3 & $96.74{\pm}0.09$ & $98.31{\pm}0.37$ & $93.68{\pm}0.18$ \\
\bottomrule
\end{tabular}}
\label{tab_ablation}
\end{table*}

\begin{figure*}[h]
\centering
\includegraphics[width=1.0\linewidth]{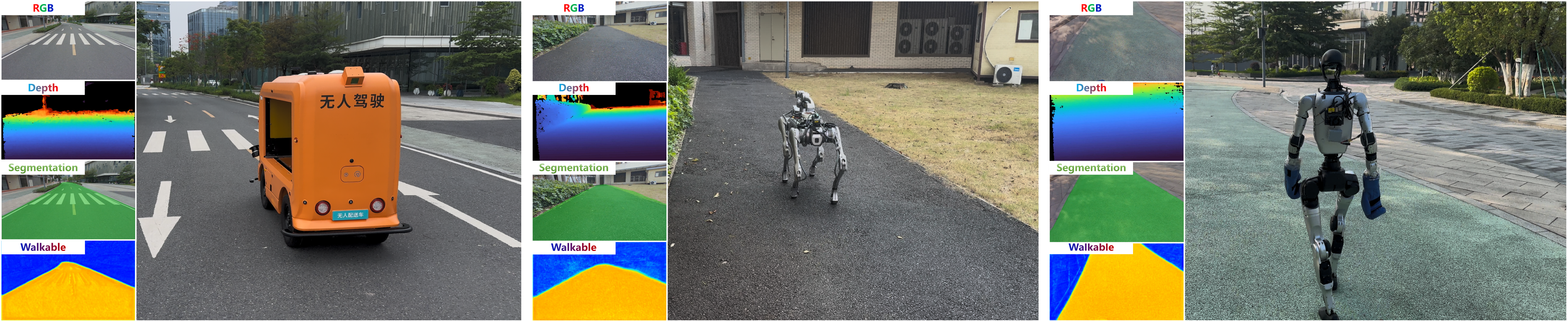}
    \caption{\textbf{\textcolor{black}{LiteViLNet RGB-and-depth demonstrations on three robot platforms.}} Left: Kuafu delivery vehicle; middle: Unitree-B2; right: Unitree-G1. Each first-person group is ordered as \textcolor{black}{RGB, aligned depth, predicted walkable-area mask, and confidence heatmap}.
    Right column shows the robot navigating autonomously using our lightweight perception system. 
    }
\label{fig:real_world_all}
\end{figure*}

\begin{figure}
    \centering
    \includegraphics[width=0.94\linewidth]{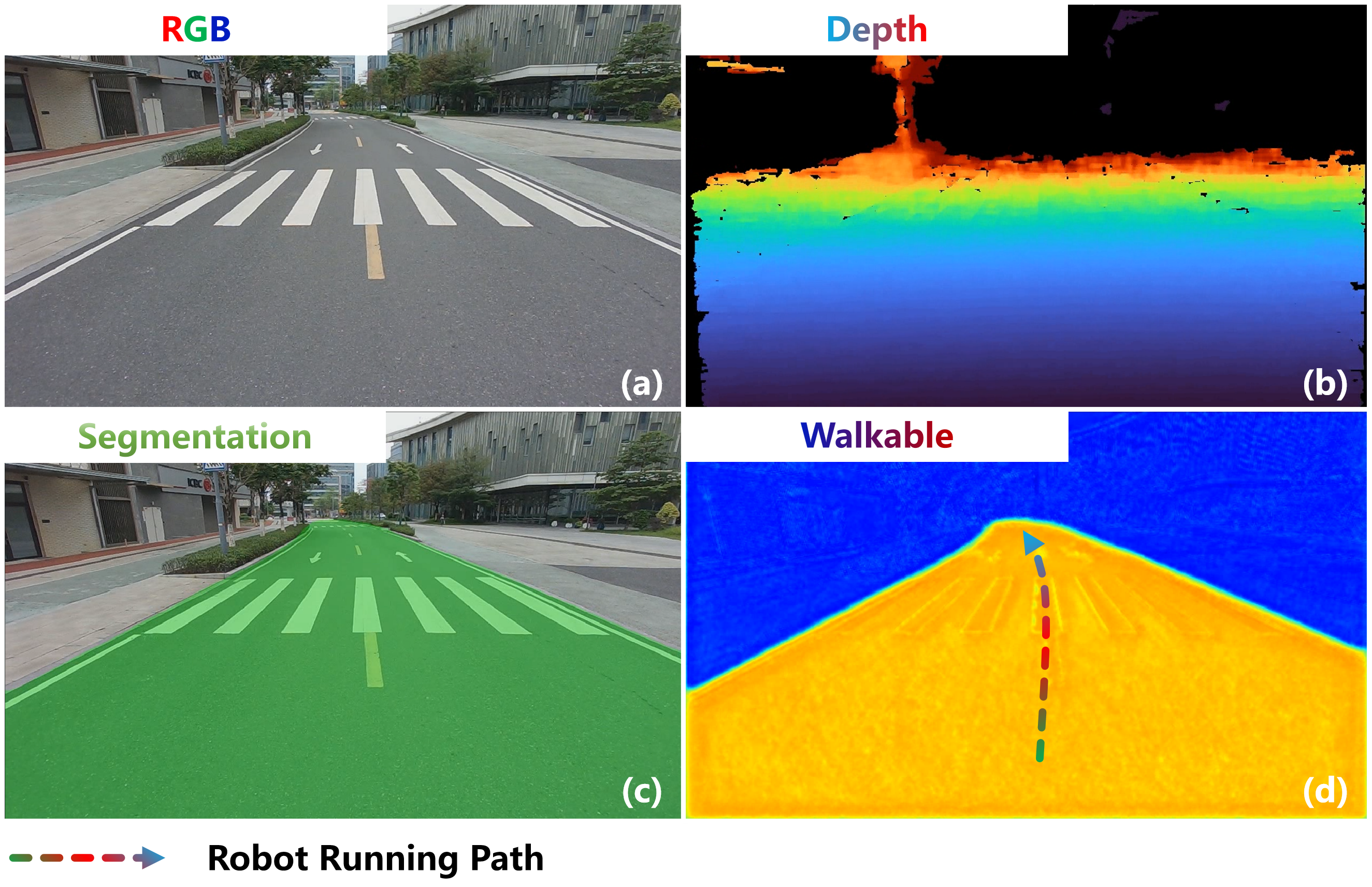}
    \caption{\textbf{First-person Perception Pipeline of LiteViLNet on the Kuafu Delivery Vehicle.
% RGB-and-depth demonstration on the Kuafu delivery vehicle.
} 
% \textcolor{black}{The panel order is (a) RGB, (b) aligned depth, (c) predicted walkable-area mask, and (d) confidence heatmap with the planned center path}.
The panels show: (a) raw RGB image from the Orbbec Gemini 336L camera, (b)
corresponding depth map, (c) drivable area segmentation mask, and (d) walkable confidence heatmap overlaid with the planned robot trajectory. The bottom legend indicates the robot running path, demonstrating that LiteViLNet provides stable and accurate road perception for lane-centering navigation in real-world scenarios.
}
    \label{fig:real_experiment_view}
\end{figure}

\subsection{Ablation Study}\label{sec:ablation}

% \textcolor{blue}{Table~\ref{tab_ablation} evaluates the cumulative design on the heterogeneous ORFD test set. The direct-addition geometry row provides the multi-modal reference for learned fusion. Replacing direct addition with MSFM improves mean F-score and IoU by 0.74 and 1.36 points, respectively, with positive same-seed gains in all three runs. The large-kernel bridge further raises all three mean metrics, with its strongest incremental gain being 0.91 points in AP.}
% \textcolor{blue}{The full model reaches $96.74\pm0.09\%$ F-score, $98.31\pm0.37\%$ AP, and $93.68\pm0.18\%$ IoU. It exceeds RGB-only in all three paired runs, with mean gains of 0.41, 0.76, and 0.76 points in F-score, AP, and IoU, respectively. Deep supervision yields the smallest F-score and IoU variation in the cumulative sequence: the corresponding sample SDs decrease from 0.55/1.02 to 0.09/0.18 while retaining high mean scores. Collectively, these ablation results verify that our extended MSFM fusion, large‑kernel bridge and deep supervision work synergistically to boost segmentation accuracy and training robustness for complex off‑road perception tasks.}
Table~\ref{tab_ablation} evaluates the individual contributions of each module on the heterogeneous ORFD held-out test set. When incorporating geometric features via simple scale-wise element-wise fusion, AP sees a mild improvement, yet F-score and IoU degrade with noticeably elevated variance. Substituting simple addition with our extended MSFM fusion restores and improves segmentation performance, lifting mean F-score and IoU by 0.74 and 1.37 points, respectively. Adding the large-kernel bridge pushes AP further to $98.44\pm0.39\%$, validating its capacity to capture long-range contextual cues.

Our full model augmented with deep supervision achieves $96.74\pm0.09\%$ F-score, $98.31\pm0.37\%$ AP, and $93.68\pm0.18\%$ IoU. Deep supervision delivers substantial stabilization across training runs: sample standard deviations for F-score and IoU drop from $0.55$/$1.02$ to $0.09$/$0.18$, while maintaining competitive mean accuracy. Collectively, the results confirm that MSFM, large-kernel bridge and deep supervision provide complementary benefits to achieve balanced segmentation performance and strong run-to-run robustness for off-road multi-modal perception.

\subsection{Real-World Experiments}\label{sec:realworld}

We additionally demonstrate perception-and-control integration on a Kuafu delivery vehicle, Unitree-B2, and Unitree-G1 (Fig.~\ref{fig:real_world_all}). The Orbbec Gemini 336L provides aligned RGB and dense depth. For this camera-derived depth input, millimeter depth $D$ is efficiently encoded as \textcolor{black}{$D_n=\operatorname{clip}(D,0,12000)/12000$, a validity mask $M=\mathbf 1[0<D<12000]$, and inverse depth $M(1-D_n)$}; the three channels are mapped to $[-1,1]$. We call this representation \textcolor{black}{\texttt{depth3}} and feed it to the same geometry stream.
The common geometry stream uses \textcolor{black}{few-shot, session-specific fine-tuning with 10-25 manually annotated frames per platform setting}, demonstrating efficient adaptation beyond the KITTI point-cloud representation. The $1280\times800$ camera is configured at 15 FPS. 
% Under the model-only $384\times1248$, batch-1, FP16, \texttt{MAXN\_SUPER} protocol, Jetson PyTorch reaches $22.18\pm0.21$ FPS and a separately built TensorRT FP16 engine reaches $68.73\pm0.06$ FPS.
% At $800\times1280$, model inference takes 107.18~ms and the complete RGB-\texttt{depth3} pipeline runs at 4.69~Hz on Jetson Orin NX, with a mean end-to-end latency of 213.19~ms including host decode and geometry construction, transfer, model inference, and mask postprocessing. Together, these measurements characterize the sensor, model-only, and end-to-end operating rates on the embedded platform.
% \textcolor{blue}{In our real-world control implementation, a center path is extracted from the predicted binary mask and converted to velocity commands, as illustrated in Fig.~\ref{fig:real_experiment_view}. These perception-control experiments demonstrate that LiteViLNet provides a common RGB-geometry walkable-area interface for a wheeled vehicle, a quadruped, and a humanoid.}
In our real‑world robotic control implementation, a central navigation path is extracted from the predicted binary walkable‑area mask and subsequently mapped into robot velocity commands, as illustrated in Fig.~\ref{fig:real_experiment_view}. These perception‑to‑control experiments demonstrate that LiteViLNet delivers a unified RGB‑geometry interface for walkable‑area estimation, which can be seamlessly deployed across distinct hardware platforms including a wheeled delivery vehicle, a quadruped robot, and a humanoid robot.

\section{Conclusion}
\textcolor{black}{LiteViLNet combines a compact RGB encoder, a 0.12M-parameter geometry encoder, multi-scale cross-modal fusion, a large-kernel bridge, and a deeply supervised decoder. It obtains $97.23\pm0.15\%$ KITTI MaxF and $96.74\pm0.09\%$ F-score/$93.68\pm0.18\%$ IoU on the held-out ORFD test set, leading the ORFD comparison in F-score, AP, and IoU. The three-seed ORFD ablation verifies consistent F-score and IoU gains from MSFM, a 0.91-point mean AP gain from the large-kernel bridge, and reduced run-to-run variation with deep supervision. The model reaches $22.18\pm0.21$ FPS on Jetson Orin NX in model-only PyTorch FP16 inference. Camera-depth perception-control experiments on three robots further demonstrate the portability of the design to embedded robotic perception.}

% \noindent\textcolor{blue}{\textbf{Future Work.} 
% Building on the shared perception-control interface demonstrated here, future work will extend LiteViLNet with robot-conditioned, multi-level traversability maps, uncertainty-aware geometric reasoning, and tighter integration with planning and control.
Based on the shared perception-control interface validated in this paper, future research will extend LiteViLNet to construct robot-conditioned multi-level traversability maps, enable uncertainty-aware geometric reasoning, and achieve closer integration with robotic planning and control subsystems.

% \normalem
% \renewcommand{\bibfont}{\small}
% \setlength{\leftmargini}{5pt}  % 调整左边界
% \setlength{\bibhang}{1.5em}    % 调整悬挂缩进

% \normalem
% \renewcommand{\bibfont}{\small}
% \setlength{\bibhang}{0pt}
% \printbibliography

\defbibenvironment{bibliography}
  {\list
     {\printtext[labelnumberwidth]{%
        \printfield{labelprefix}%
        \printfield{labelnumber}}}
     {\setlength{\labelwidth}{\labelnumberwidth}%
      \setlength{\leftmargin}{\labelwidth}%   % 让左边距等于编号宽度
      \addtolength{\leftmargin}{\labelsep}%    % 加上编号与文字之间的间距
      \setlength{\itemsep}{\bibitemsep}%       % 引用条目间的距离
      \setlength{\parsep}{\bibparsep}}%
      \renewcommand*{\makelabel}[1]{##1\hss}} % 关键：##1\hss 确保编号在左侧对齐
  {\endlist}
  {\item}

\normalem
\renewcommand{\bibfont}{\fontsize{8}{7.8}\selectfont}
\setlength{\labelsep}{0.25em}  % 标号和内容之间的间距
\printbibliography

@article{early_road,
  title={Road Detection Based on Illuminant Invariance},
  author={{\'A}lvarez, Jos{\'e} M. and L{\'o}pez, Antonio M.},
  journal={IEEE Transactions on Intelligent Transportation Systems},
  volume={12},
  number={1},
  pages={184--193},
  year={2011},
  publisher={IEEE},
  doi={10.1109/TITS.2010.2076349}
}

@inproceedings{nadeem2026lightweight,
  title={{PIDNet}: A Real-Time Semantic Segmentation Network Inspired by {PID} Controllers},
  author={Xu, Jiacong and Xiong, Zixiang and Bhattacharyya, Shankar P},
  booktitle={Proceedings of the IEEE/CVF Conference on Computer Vision and Pattern Recognition},
  pages={19529--19539},
  year={2023},
  doi={10.1109/CVPR52729.2023.01871},
  eprint={2206.02066},
  archivePrefix={arXiv},
  url={https://arxiv.org/abs/2206.02066}
}

@article{ahmed2025twinlitenet,
  title={{TwinLiteNet+}: An Enhanced Multi-Task Segmentation Model for Autonomous Driving},
  author={Che, Quang-Huy and Le, Duc-Tri and Pham, Minh-Quan and Nguyen, Vinh-Tiep and Lam, Duc-Khai},
  journal={Computers and Electrical Engineering},
  volume={128},
  pages={110694},
  year={2025},
  doi={10.1016/j.compeleceng.2025.110694}
}

@article{li2025knowledge,
  title={Knowledge Generation and Distillation for Road Segmentation in Intelligent Transportation Systems},
  author={Wang, Jianyong and Gao, Mingliang and Zhai, Wenzhe and Rida, Imad and Zhu, Xianxun and Li, Qilei},
  journal={IEEE Transactions on Intelligent Transportation Systems},
  pages={1--13},
  year={2025},
  doi={10.1109/TITS.2025.3577794}
}

@article{wang2025sdfnet,
  title={{SDFNet} for Real-Time Semantic Segmentation on Urban Road Images},
  author={Cao, Yi and Qu, Hongbo},
  journal={IAENG International Journal of Computer Science},
  volume={52},
  number={12},
  pages={4815--4821},
  year={2025},
  url={https://www.iaeng.org/IJCS/issues_v52/issue_12/IJCS_52_12_25.pdf}
}

@article{zhang2025lcirenet,
  title={{LCIRE-Net}: Lightweight Cross-Modal Information Interaction for Road Feature Extraction From Remote Sensing Images and {GPS} Trajectory/{LiDAR}},
  author={Duan, Yifei and Yang, Dan and Qu, Xiaochen and Zhang, Le and Chao, Lu and Gan, Peilu and Yuan, Shuai and Qin, Hanlin and Qu, Junsuo},
  journal={IEEE Transactions on Geoscience and Remote Sensing},
  volume={63},
  pages={1--18},
  year={2025},
  doi={10.1109/TGRS.2024.3516840}
}

@inproceedings{ma2025annotation,
  title={Annotation-free curb detection leveraging altitude difference image},
  author={Ma, Fulong and Hou, Peng and Liu, Yuxuan and Liu, Yang and Liu, Ming and Ma, Jun},
  booktitle={2025 IEEE/RSJ International Conference on Intelligent Robots and Systems (IROS)},
  pages={762--768},
  year={2025},
  organization={IEEE},
  doi={10.1109/IROS60139.2025.11247167}
}

@inproceedings{liu2021swin,
  title={{Swin Transformer}: Hierarchical Vision Transformer Using Shifted Windows},
  author={Liu, Ze and Lin, Yutong and Cao, Yue and Hu, Han and Wei, Yixuan and Zhang, Zheng and Lin, Stephen and Guo, Baining},
  booktitle={Proceedings of the IEEE/CVF International Conference on Computer Vision},
  pages={9992--10002},
  year={2021},
  doi={10.1109/ICCV48922.2021.00986}
}

@inproceedings{howard2019searching,
  title={Searching for {MobileNetV3}},
  author={Howard, Andrew and Sandler, Mark and Chu, Grace and Chen, Liang-Chieh and Chen, Bo and Tan, Mingxing and Wang, Weijun and Zhu, Yukun and Pang, Ruoming and Vasudevan, Vijay and Le, Quoc V. and Adam, Hartwig},
  booktitle={Proceedings of the IEEE/CVF International Conference on Computer Vision},
  pages={1314--1324},
  year={2019},
  doi={10.1109/ICCV.2019.00140},
  eprint={1905.02244},
  archivePrefix={arXiv},
  url={https://arxiv.org/abs/1905.02244}
}

@article{cmx2022,
  title={{CMX}: Cross-Modal Fusion for {RGB-X} Semantic Segmentation With Transformers},
  author={Zhang, Jiaming and Liu, Huayao and Yang, Kailun and Hu, Xinxin and Liu, Ruiping and Stiefelhagen, Rainer},
  journal={IEEE Transactions on Intelligent Transportation Systems},
  volume={24},
  number={12},
  pages={14679--14694},
  year={2023},
  publisher={IEEE},
  doi={10.1109/TITS.2023.3300537}
}

@article{snerv2,
  title={{SNE-RoadSegV2}: Advancing Heterogeneous Feature Fusion and Fallibility Awareness for Freespace Detection},
  author={Feng, Yi and Ma, Yu and Andreev, Stepan and Chen, Qijun and Dvorkovich, Alexander and Pitas, Ioannis and Fan, Rui},
  journal={IEEE Transactions on Instrumentation and Measurement},
  volume={74},
  pages={1--9},
  year={2025},
  publisher={IEEE},
  doi={10.1109/TIM.2025.3545498}
}

@article{roadformer,
  title={{RoadFormer}: Duplex Transformer for {RGB-Normal} Semantic Road Scene Parsing},
  author={Li, Jiahang and Zhang, Yikang and Yun, Peng and Zhou, Guangliang and Chen, Qijun and Fan, Rui},
  journal={IEEE Transactions on Intelligent Vehicles},
  volume={9},
  number={7},
  pages={5163--5172},
  year={2024},
  publisher={IEEE},
  doi={10.1109/TIV.2024.3388726}
}

@article{lrdnet,
  title={{LRDNet}: Lightweight {LiDAR} Aided Cascaded Feature Pools for Free Road Space Detection},
  author={Khan, Abdullah Aman and Shao, Jie and Rao, Yunbo and She, Lei and Shen, Heng Tao},
  journal={IEEE Transactions on Multimedia},
  volume={27},
  pages={652--664},
  year={2025},
  publisher={IEEE},
  doi={10.1109/TMM.2022.3230330}
}

@inproceedings{usnet,
  title={Fast Road Segmentation via Uncertainty-aware Symmetric Network},
  author={Chang, Yicong and Xue, Feng and Sheng, Fei and Liang, Wenteng and Ming, Anlong},
  booktitle={2022 IEEE International Conference on Robotics and Automation (ICRA)},
  pages={11124--11130},
  year={2022},
  organization={IEEE},
  doi={10.1109/ICRA46639.2022.9812452}
}

@inproceedings{long2015fully,
  title={Fully convolutional networks for semantic segmentation},
  author={Long, Jonathan and Shelhamer, Evan and Darrell, Trevor},
  booktitle={Proceedings of the IEEE Conference on Computer Vision and Pattern Recognition},
  pages={3431--3440},
  year={2015},
  doi={10.1109/CVPR.2015.7298965}
}

@inproceedings{ronneberger2015u,
  title={{U-Net}: Convolutional Networks for Biomedical Image Segmentation},
  author={Ronneberger, Olaf and Fischer, Philipp and Brox, Thomas},
  booktitle={International Conference on Medical Image Computing and Computer-Assisted Intervention},
  pages={234--241},
  year={2015},
  organization={Springer},
  doi={10.1007/978-3-319-24574-4_28}
}

@inproceedings{sneroad,
  title={{SNE-RoadSeg}: Incorporating Surface Normal Information into Semantic Segmentation for Accurate Freespace Detection},
  author={Fan, Rui and Wang, Hengli and Cai, Peide and Liu, Ming},
  booktitle={European Conference on Computer Vision},
  pages={340--356},
  year={2020},
  organization={Springer},
  doi={10.1007/978-3-030-58577-8_21}
}

@inproceedings{early_fusion,
  title={Early Fusion of Camera and {LiDAR} for Robust Road Detection Based on {U-Net} {FCN}},
  author={Wulff, Florian and Sch{\"a}ufele, Bernd and Sawade, Oliver and Becker, Daniel and Henke, Birgit and Radusch, Ilja},
  booktitle={2018 IEEE Intelligent Vehicles Symposium (IV)},
  pages={1426--1431},
  year={2018},
  organization={IEEE},
  doi={10.1109/IVS.2018.8500549}
}

@inproceedings{crossview,
  title={Cross-View Transformers for Real-Time Map-View Semantic Segmentation},
  author={Zhou, Brady and Kr{\"a}henb{\"u}hl, Philipp},
  booktitle={Proceedings of the IEEE/CVF Conference on Computer Vision and Pattern Recognition},
  pages={13750--13759},
  year={2022},
  doi={10.1109/CVPR52688.2022.01339}
}

@article{howard2017mobilenets,
  title={{MobileNets}: Efficient Convolutional Neural Networks for Mobile Vision Applications},
  author={Howard, Andrew G and Zhu, Menglong and Chen, Bo and Kalenichenko, Dmitry and Wang, Weijun and Weyand, Tobias and Andreetto, Marco and Adam, Hartwig},
  journal={arXiv preprint arXiv:1704.04861},
  year={2017},
  eprint={1704.04861},
  archivePrefix={arXiv},
  url={https://arxiv.org/abs/1704.04861}
}

@inproceedings{zhang2018shufflenet,
  title={{ShuffleNet V2}: Practical Guidelines for Efficient {CNN} Architecture Design},
  author={Ma, Ningning and Zhang, Xiangyu and Zheng, Hai-Tao and Sun, Jian},
  booktitle={Proceedings of the European Conference on Computer Vision (ECCV)},
  pages={122--138},
  year={2018},
  doi={10.1007/978-3-030-01264-9_8}
}

@inproceedings{mehta2018espnet,
  title={{ESPNet}: Efficient Spatial Pyramid of Dilated Convolutions for Semantic Segmentation},
  author={Mehta, Sachin and Rastegari, Mohammad and Caspi, Anat and Shapiro, Linda and Hajishirzi, Hannaneh},
  booktitle={Proceedings of the European Conference on Computer Vision (ECCV)},
  pages={561--580},
  year={2018},
  doi={10.1007/978-3-030-01249-6_34}
}

@inproceedings{yu2018bisenet,
  title={{BiSeNet}: Bilateral Segmentation Network for Real-Time Semantic Segmentation},
  author={Yu, Changqian and Wang, Jingbo and Peng, Chao and Gao, Changxin and Yu, Gang and Sang, Nong},
  booktitle={Proceedings of the European Conference on Computer Vision (ECCV)},
  pages={334--349},
  year={2018},
  doi={10.1007/978-3-030-01261-8_20}
}

@inproceedings{eca,
  title={{ECA-Net}: Efficient Channel Attention for Deep Convolutional Neural Networks},
  author={Wang, Qilong and Wu, Banggu and Zhu, Pengfei and Li, Peihua and Zuo, Wangmeng and Hu, Qinghua},
  booktitle={Proceedings of the IEEE/CVF Conference on Computer Vision and Pattern Recognition},
  pages={11531--11539},
  year={2020},
  doi={10.1109/CVPR42600.2020.01155}
}

@inproceedings{coordinate,
  title={Coordinate Attention for Efficient Mobile Network Design},
  author={Hou, Qibin and Zhou, Daquan and Feng, Jiashi},
  booktitle={Proceedings of the IEEE/CVF Conference on Computer Vision and Pattern Recognition},
  pages={13708--13717},
  year={2021},
  doi={10.1109/CVPR46437.2021.01350}
}

@article{kitti,
  title={Vision Meets Robotics: The {KITTI} Dataset},
  author={Geiger, Andreas and Lenz, Philip and Stiller, Christoph and Urtasun, Raquel},
  journal={The International Journal of Robotics Research},
  volume={32},
  number={11},
  pages={1231--1237},
  year={2013},
  publisher={Sage Publications Sage UK: London, England},
  doi={10.1177/0278364913491297}
}

@inproceedings{sun2025rod,
  title={{ROD}: {RGB}-Only Fast and Efficient Off-Road Freespace Detection},
  author={Sun, Tong and Ye, Hongliang and Mei, Jilin and Chen, Liang and Zhao, Fangzhou and Zong, Leiqiang and Hu, Yu},
  booktitle={2025 IEEE International Conference on Robotics and Automation (ICRA)},
  pages={9787--9793},
  year={2025},
  organization={IEEE},
  doi={10.1109/ICRA55743.2025.11127491}
}

@inproceedings{milioto2019rangenet++,
  title={{RangeNet++}: Fast and Accurate {LiDAR} Semantic Segmentation},
  author={Milioto, Andres and Vizzo, Ignacio and Behley, Jens and Stachniss, Cyrill},
  booktitle={2019 IEEE/RSJ International Conference on Intelligent Robots and Systems (IROS)},
  pages={4213--4220},
  year={2019},
  organization={IEEE},
  doi={10.1109/IROS40897.2019.8967762}
}

@article{ma2026annotation,
  title={Annotation-Free Detection of Drivable Areas and Curbs Leveraging {LiDAR} Point Cloud Maps},
  author={Ma, Fulong and Peng, Daojie and Ma, Jun},
  journal={arXiv preprint arXiv:2603.27553},
  year={2026},
  eprint={2603.27553},
  archivePrefix={arXiv},
  url={https://arxiv.org/abs/2603.27553}
}

@inproceedings{min2022orfd,
  title={{ORFD}: A Dataset and Benchmark for Off-Road Freespace Detection},
  author={Min, Chen and Jiang, Weizhong and Zhao, Dawei and Xu, Jiaolong and Xiao, Liang and Nie, Yiming and Dai, Bin},
  booktitle={2022 IEEE International Conference on Robotics and Automation (ICRA)},
  pages={2532--2538},
  year={2022},
  organization={IEEE},
  doi={10.1109/ICRA46639.2022.9812139},
  eprint={2206.09907},
  archivePrefix={arXiv},
  url={https://arxiv.org/abs/2206.09907}
}

@inproceedings{ma2023self,
  title={Self-Supervised Drivable Area Segmentation Using {LiDAR}'s Depth Information for Autonomous Driving},
  author={Ma, Fulong and Liu, Yang and Wang, Sheng and Wu, Jin and Qi, Weiqing and Liu, Ming},
  booktitle={2023 IEEE/RSJ International Conference on Intelligent Robots and Systems (IROS)},
  pages={41--48},
  year={2023},
  organization={IEEE},
  doi={10.1109/IROS55552.2023.10341687}
}

@article{chen2019progressive,
  title={Progressive {LiDAR} adaptation for road detection},
  author={Chen, Zhe and Zhang, Jing and Tao, Dacheng},
  journal={IEEE/CAA Journal of Automatica Sinica},
  volume={6},
  number={3},
  pages={693--702},
  year={2019},
  publisher={IEEE},
  doi={10.1109/JAS.2019.1911459}
}

@article{zhao2025curbnet,
  title={{CurbNet}: Curb Detection Framework Based on {LiDAR} Point Cloud Segmentation},
  author={Zhao, Guoyang and Ma, Fulong and Qi, Weiqing and Liu, Yuxuan and Liu, Ming and Ma, Jun},
  journal={IEEE Transactions on Intelligent Transportation Systems},
  volume={26},
  number={6},
  pages={8961--8974},
  year={2025},
  publisher={IEEE},
  doi={10.1109/TITS.2025.3540645}
}

@article{peng2025lovon,
  title={{LOVON}: Legged Open-Vocabulary Object Navigator},
  author={Peng, Daojie and Cao, Jiahang and Zhang, Qiang and Ma, Jun},
  journal={arXiv preprint arXiv:2507.06747},
  year={2025},
  eprint={2507.06747},
  archivePrefix={arXiv},
  url={https://arxiv.org/abs/2507.06747}
}

\end{document}